\documentclass[conference]{IEEEtran}
\IEEEoverridecommandlockouts
\usepackage{cite}
\usepackage{amsmath,amssymb,amsfonts}
\usepackage{algorithmic}
\usepackage{threeparttable}
\usepackage{multirow}
\usepackage{array}
\newcolumntype{u}[1]{>{\centering\arraybackslash}p{#1}}
\usepackage{comment}
\usepackage[table]{xcolor}
\usepackage{graphicx}
\usepackage{textcomp}
\usepackage{xcolor}
\usepackage{bigints}
\usepackage{amssymb}
\usepackage{url}
\usepackage[caption=false]{subfig}
\usepackage{tabu}
\usepackage{makecell}
\def\BibTeX{{\rm B\kern-.05em{\sc i\kern-.025em b}\kern-.08em
    T\kern-.1667em\lower.7ex\hbox{E}\kern-.125emX}}
    
{}
\newcommand{\napps}{\text{{8}}}{}
\newcommand{\ineq}[1]{\footnotesize$#1$\normalsize}{}

\definecolor{azuremist}{rgb}{0.94, 1.0, 1.0}
\definecolor{babyblueeyes}{rgb}{0.63, 0.79, 0.95}
\definecolor{beaublue}{rgb}{0.74, 0.83, 0.9}
\definecolor{beige}{rgb}{0.96, 0.96, 0.86}
\definecolor{bisque}{rgb}{1.0, 0.89, 0.77}
\definecolor{gainsboro}{rgb}{0.86, 0.86, 0.86}
\definecolor{ghostwhite}{rgb}{0.97, 0.97, 1.0}

\begin{document}
\bstctlcite{IEEEexample:BSTcontrol}

\title{
%Learning in Recurrent Spiking Neural Networks using full-FORCE Training
Learning in Feedback-driven Recurrent Spiking Neural Networks using full-FORCE Training
%Learning in Generalized Recurrent Spiking Neural Networks with Delayed Feedback Loop
}

\author{\IEEEauthorblockN{Ankita Paul, Stefan Wagner and Anup Das}
\IEEEauthorblockA{{Drexel University}\\
Philadelphia, USA \\
\{ankita.paul,saw368,anup.das\}@drexel.edu}
}
%\author{}

\maketitle

\begin{abstract}
Feedback-driven recurrent spiking neural networks (RSNNs) are powerful computational models that can mimic dynamic systems.
However, the presence of a feedback loop from the readout to the recurrent layer de-stabilizes the learning mechanism and prevents it from converging.
Here, we propose a supervised training procedure for RSNNs, where a second network is introduced \textit{only} during the training, to provide hint for the target dynamics.
%Recurrent Spiking Neural Networks (RSNNs) are powerful computation models that enable spatio-temporal learning to mimic dynamic systems and facilitate energy-efficient implementation on event-driven neuromorphic hardware platforms. However, due to the delayed feedback loop, RSNNs suffer from the training convergence issue, in addition to their high computational complexity of unfolding in time, vanishing gradient, and approximate differentiation.
%We propose a supervised training procedure for RSNNs, where a second network is introduced during training to provide hints for the target dynamics. 
%Contrary to existing training approaches where only the output layer of an RSNN is learned, 
The proposed training procedure consists of generating targets for both recurrent and readout layers (i.e., for a full RSNN system).
It uses the recursive least square-based First-Order and Reduced Control Error (FORCE) algorithm to fit the activity of each layer to its target. 
The proposed full-FORCE training procedure reduces the amount of modifications needed to keep the error between the output and target close to zero.
These modifications control the feedback loop, which causes the training to converge.
We demonstrate the improved performance and noise robustness of the proposed full-FORCE training procedure to model \napps{} dynamical systems using RSNNs with leaky integrate and fire (LIF) neurons and rate coding. 
For energy-efficient hardware implementation, an alternative time-to-first-spike (TTFS) coding is implemented for the full-FORCE training procedure. 
Compared to rate coding, full-FORCE with TTFS coding generates fewer spikes and facilitates faster convergence to the target dynamics.
% with only marginally lower performance.
\end{abstract}


\section{Introduction}\label{sec:introduction}

Spiking Neural Networks (SNNs) are biologically-realistic computational models that exploit the dynamics of the central nervous system in primates~\cite{maass1997networks}. 
SNNs facilitate energy-efficient VLSI implementation on an event-driven neuromorphic hardware platform such as DYNAPs~\cite{dynapse}, $\mu$Brain~\cite{sentryos}, and Loihi~\cite{loihi}.
%, and TrueNorth~\cite{truenorth}.
Recurrent spiking neural networks (RSNNs) are a type of SNNs that can accurately mimic a dynamical system~\cite{voelker2019dynamical}, where mimicking involves modeling the interaction of system components over time and performing analogous tasks when inputs are silent over thousands of time steps~\cite{han2004modeling}.

Reservoir computing is 
an important example of RSNNs~\cite{lukovsevivcius2009reservoir,gauthier2021next,tanaka2019recent}.
It consists of an input layer, a reservoir layer of recurrently-connected neurons, and an output readout layer (see Figure~\ref{fig:rsnns}).
The reservoir may receive feedback from the readout layer~\cite{li2012selective}.
Training a reservoir computing system is challenging for three reasons~\cite{bellec2020solution,bengio1994learning,pascanu2013difficulty}.
First, the feedback loop can prevent the training to converge. This is due to the delayed effect from the feedback signal, which can cause the output to diverge away from the target response~\cite{chung2015gated}.
Second, it suffers from the credit assignment problem of identifying which neurons and synapses have the highest impact on the output~\cite{bengio2013advances}.
Finally, RSNNs that have spontaneous chaotic activity, i.e., systems that are active without any input in the pre-training phase 
are difficult to train~\cite{rajan2010stimulus,sussillo2009generating,soula2006spontaneous,engelken2020lyapunov}.
%prior to training, do not converge during training.

To address these limitations, prior learning approaches have introduced several simplifications such as 1) eliminating the feedback loop, 2) training only the synaptic connections between the reservoir and readout, and 3) suppressing chaotic activity in the pre-training phase.
The Liquid State Machine (LSM) is one such reservoir computing system with the aforementioned simplifications~\cite{lsm,HeartEstmNN,grzyb2009model}.

Some recent efforts have addressed the training of feedback-driven RSNNs that may include spontaneous chaotic activity in the pre-training phase. In~\cite{bptt}, backpropagation through time (BPTT) is proposed for SNNs, which is extended in~\cite{zhang2019spike,zhang2020temporal,zhang2021spiking} to train both the reservoir and readout of an LSM. 
In these approaches, the reservoir is trained using an unsupervised algorithm such as spike timing dependent plasticity (STDP)~\cite{stdp}, while the readout is trained using BPTT.
%(no feedback loop).
Following are its limitations.
%First, BPTT requires 
%Apart from the necessity of 
%approximate differentiation of discrete spike events, which leads to a lower accuracy.
%training LSMs using BPTT presents several additional challenges. 
First, BPTT requires unfolding an RSNN into effective network layers, one for each time stamp and obtaining gradient over the time period. 
This 
%unfolding in time 
leads to an exponential increase in the computational complexity~\cite{zhang2019spike}.
Second, BPTT suffers from the vanishing gradient problem, where certain components of the gradient computed on the loss function may get close to zero~\cite{hochreiter1998vanishing}.
Vanishing gradients stall parameter updates during 
the training.

In~\cite{force_snn}, a recursive least square-based First-Order and Reduced Control Error (FORCE) algorithm is proposed to train the readout layer of an RSNN with a feedback signal from the readout to the spiking neurons in the reservoir.
Contrary to conventional training procedures, where weight modifications are performed to reduce large errors in the output, objective of the FORCE training procedure is to reduce the amount of modifications needed to keep the error small.
Such modifications are then used to control the feedback signal without needing any additional training mechanisms for the feedback loop.
FORCE training procedure can train networks that exhibit chaotic activity in the pre-training phase.
However, FORCE training suffers from the following limitations. First, FORCE can \textit{only} model dynamical systems that can be represented using a closed-form differential equation~\cite{sussillo2009generating}. However, there are many practical systems where such closed-form representation is difficult to obtain~\cite{zimmermann2002modeling}.
Second, FORCE training procedure is susceptible to input noise~\cite{lim2021noisy}. This is because FORCE trains only the readout and not the reservoir. Therefore, noise introduces a small variation in the response of the reservoir neurons, which triggers a domino effect using the feedback loop and eventually, diverging the output significantly from the target.

A recent training procedure, called \textbf{full-FORCE} can address the aforementioned limitations\cite{full_force}.
Here,
%To address these limitations, a supervised training procedure called \textbf{full-FORCE} is proposed, where 
a second recurrent network is introduced \textit{only} during training to obtain targets 
for the reservoir neurons.
It
%not only for the output layer but also for the reservoir layer, and then using 
uses the FORCE algorithm to fit the response of each layer (recurrent and readout) to its target response. However, full-FORCE training requires sigmoid neurons, which are not supported on event-driven neuromorphic hardware platforms.
Table~\ref{tab:summary_of_difference} summarizes these existing training procedures.

\vspace{-10pt}
\begin{table}[h!]
	\renewcommand{\arraystretch}{1.2}
	\setlength{\tabcolsep}{1.2pt}
	\caption{Training procedures for Feedback-driven RSNNs.}
	\label{tab:summary_of_difference}
	%\vspace{-10pt}
	\centering
	\begin{threeparttable}
	{\fontsize{8}{10}\selectfont
	    %\vspace{-10pt}
		\begin{tabular}{c u{1cm} u{2cm} u{1.5cm} u{1cm} u{2cm}}
		    %\tabucline[2pt]{-}
			%\hline
			\Xhline{4\arrayrulewidth}
			& \multicolumn{4}{c}{\textbf{Feedback-driven RSNNs}} &  \\
			\cline{2-5}
			& \textbf{Feedback Driven} & \textbf{Training Procedure} & \textbf{Spontaneous Activity} & \textbf{LIF Neuron} & \textbf{Neuromorphic Computing}\\
			%& \textbf{Driven}   & \textbf{Procedure} & \textbf{Activity} & \textbf{Neuron} & \\
			%\hline
			%\hline
			\Xhline{4\arrayrulewidth}
			%\tabucline[2pt]{-}
			\cite{lsm} & $\times$ & Readout Only (supervised) & $\times$ & $\surd$ & $\surd$\\
			\rowcolor{gainsboro} \cite{zhang2019spike} & $\times$ & Reservoir and Readout (BPTT) & $\times$ & $\surd$ & $\times$\\
			\cite{sussillo2009generating} & $\surd$ & Readout (FORCE) & $\surd$ & $\times$ & $\times$\\
			\rowcolor{gainsboro} \cite{full_force} & $\surd$ & Reservoir and Readout (full-FORCE) & $\surd$ & $\times$ & $\times$\\
			\cite{force_snn} & $\surd$ & Readout (FORCE) & $\surd$ & $\surd$ & $\times$\\
			\hline
			%\tabucline[2pt]{-}
			\rowcolor{azuremist} \textbf{ours} & $\surd$ & Reservoir and Readout (full-FORCE) & $\surd$ & $\surd$ & $\surd$\\
			%\hline
			\Xhline{4\arrayrulewidth}
			%\tabucline[2pt]{-}
	\end{tabular}}
	\end{threeparttable}
	%\vspace{12pt}
	%\vspace{-5pt}
\end{table}

We propose a full-FORCE implementation for feedback-driven RSNNs using the leaky integrate and fire (LIF) neurons.
We make the following key contributions.
\begin{itemize}
    \item We implement the full-FORCE algorithm for rate coding and explore the design space of network architecture for faster convergence to the target and facilitate energy-efficient neuromorphic implementation.
    \item By training both reservoir and the readout layers, we illustrate the stability of the training procedure to noise.
    \item We provide an alternative time-to-first-spike (TTFS) implementation of the full-FORCE training procedure. We show that full-FORCE with TTFS encoding generates fewer spikes and facilitates faster convergence to the target dynamics.%the target response with only marginally lower performance.
\end{itemize}

We evaluate our full-FORCE training procedure (which trains both reservoir and the readout layers) against the FORCE learning (which trains only the readout layer).
%, where only the output layer of an RSNN is trained to produce the target response. Although both these training procedures are designed for LIF neurons, 
%with rate encoding which make them suitable for implementation on a neuromorphic hardware, 
Results demonstrate that our full-FORCE implementation has an average 51\% lower mean square error (MSE) and it provides higher tolerance to noise. Furthermore, the proposed full-FORCE implementation with TTFS coding results in {50\%} fewer spikes and {40\%} lower convergence time.%, with only \textcolor{red}{z\%} lower performance.

\section{Preliminaries}\label{sec:prelim}
\begin{figure}[h!]
	\centering
	%\vspace{-10pt}
	\centerline{\includegraphics[width=0.99\columnwidth]{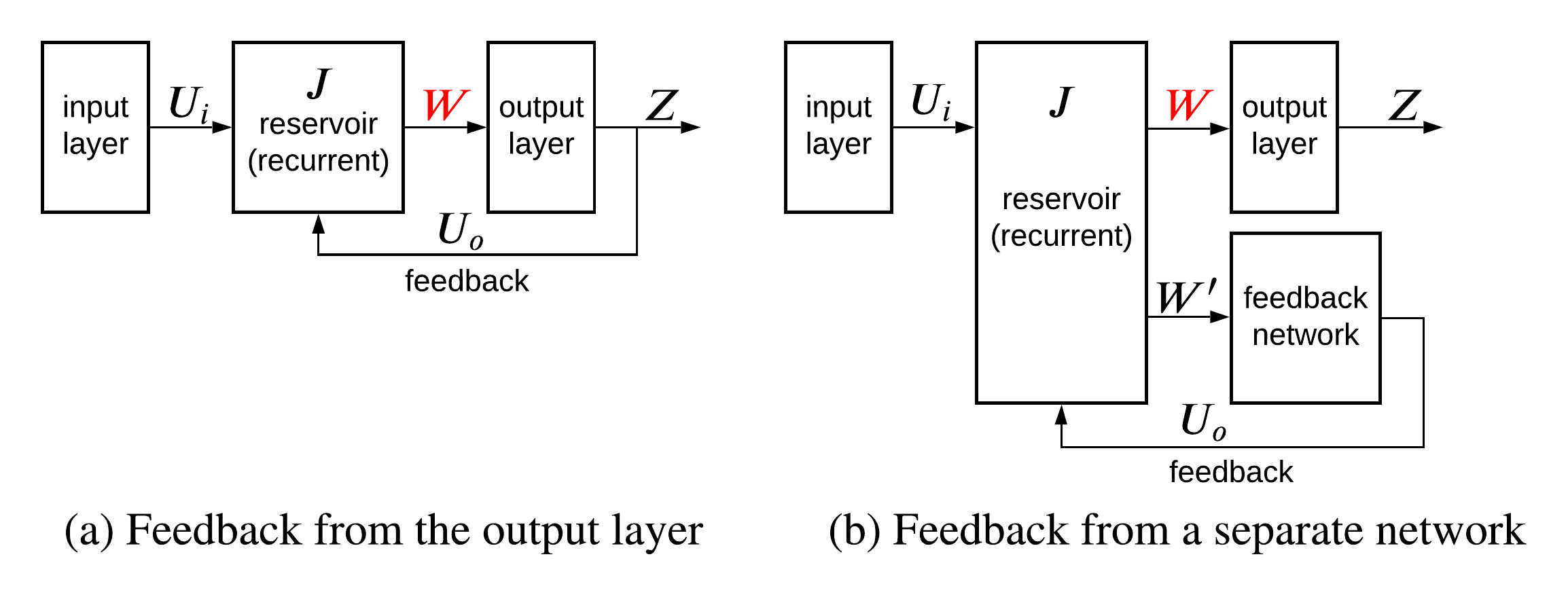}}
	%\vspace{-10pt}
	%\caption{An example of spiking neural network.}
	\caption{Different reservoir computing architectures. 
	%Connections that are modified during training are indicated in red. 
	(a) Feedback to the recurrent reservoir is provided by the output readout units. 
	(b) Feedback to the reservoir is provided by a separate feedback network.}
	\label{fig:rsnns}
\end{figure}

Figure~\ref{fig:rsnns} shows two reservoir computing architectures~\cite{lukovsevivcius2009reservoir}. 
Both these architectures consist of an input layer, a reservoir layer of recurrently-connected neurons, and an output readout layer. 
The reservoir may also receive feedback from the readout.
In Figure~\ref{fig:rsnns}a,
the feedback is generated directly from the readout layer.
In Figure~\ref{fig:rsnns}b, the feedback is generated from a separate (recurrent) network, which can be trained offline.
%Typically, training of RSNNs involve updating the synaptic connections between the reservoir and the output readout layer as showing in red in Figure~\ref{fig:rsnns}.

\subsection{Difficulty in Training with Feedback Loop} Training a reservoir computing architecture with feedback is challenging. To understand this, we observe that there are two pathways in this architecture. The direct pathway is from the reservoir to the readout layer, while
%generating the output response \ineq{Z}. The 
the feedback pathway is from the readout to the reservoir layer.
Considering only the direct pathway, it is straightforward to optimize the weights \ineq{W} such that the output \ineq{Z(t)} is close to the target response of a dynamical system.
However, due to the delay on the feedback pathway, the effect of weight updates on the output \ineq{Z(t)} arrives at the reservoir with a time lag.
Training the weights via the direct pathway reduces the output error.
However, the delayed feedback from the readout causes the output \ineq{Z} to deviate away from the target, increasing the error.

\begin{figure*}[h!]
	\centering
	%\vspace{-10pt}
	\centerline{\includegraphics[width=1.99\columnwidth]{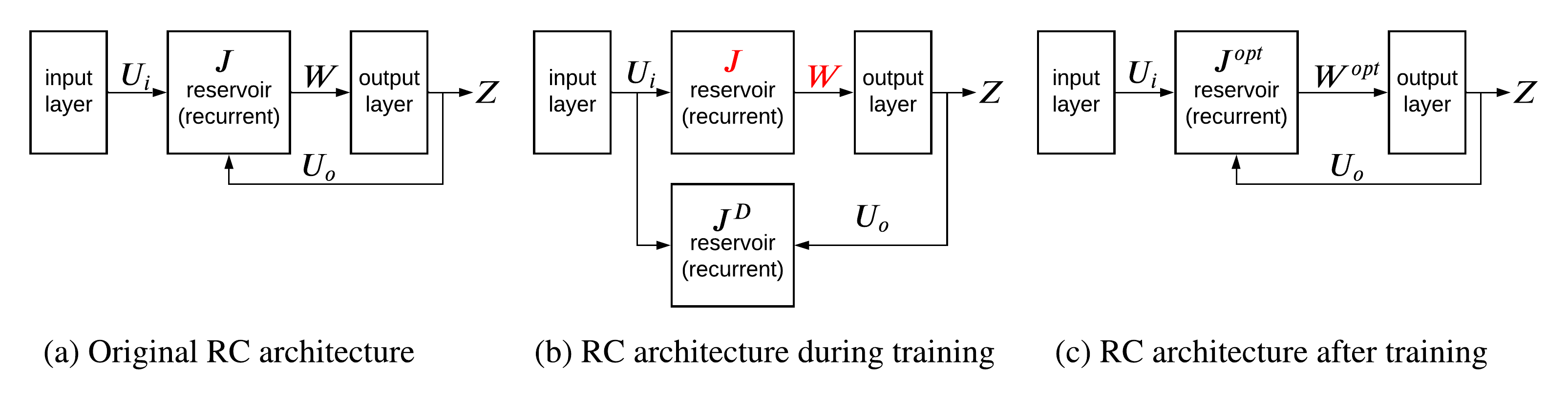}}
	%\vspace{-10pt}
	%\caption{An example of spiking neural network.}
	\caption{Learning in reservoir computing (RC) architecture using the full-FORCE training. (a) Original RC architecture. (b) Training an RC architecture using a separate target network. (c) A fully trained RC architecture.}
	\label{fig:full_force}
\end{figure*}

\subsection{Training the Readout using FORCE}\label{sec:force}
To introduce the FORCE training procedure, we focus on the architecture of Figure~\ref{fig:rsnns}a, which can then be generalized to Figure~\ref{fig:rsnns}b.
We elaborate the training procedure considering sigmoidal neurons (see Section~\ref{sec:spiking_conversion} for RSNNs).

Let the architecture of Figure~\ref{fig:rsnns}a be designed to model a dynamical system whose response at time \ineq{t} is \ineq{f(t)}. We introduce the following notations.

\begin{footnotesize}
\begin{align*}
%{M} =&~\text{Input dimension}\\
%{N} =&~\text{Number of neurons in the reservoir}\\\
%{P} =&~\text{Dimension of the readout}\\
f(t) =&~\text{Input and target}\\
U_i =&~\text{Synaptic weights between input and reservoir.}\\
J =&~\text{Synaptic weights between the reservoir neurons.}\\
Y(t) =&~\text{Output of the reservoir.}\\
W(t) =&~\text{Synaptic weights between reservoir and readout.}\\
U_o(t) =&~\text{Synaptic weights between readout and reservoir (feedback).}\\
Z(t) =&~\text{Output of the readout.}
\end{align*}
\end{footnotesize}
\normalsize

%Let the input dimension be \ineq{m} inputs, \ineq{n} 
%Let the matrix \ineq{U_i\in \mathbb{R}} defines the connection of the input layer neurons to the reservoir neurons, \ineq{J} defines the internal connections of the reservoir neurons, \ineq{Y(t)} be the response of the reservoir neurons, \ineq{W (t)} defines the connection of reservoir neurons to the readout layer neurons, and \ineq{U_o} defines the connection on the feedback path.
Output of the architecture at time \ineq{t} is
\begin{equation}
    \label{eq:output}
    \footnotesize Z(t) = W^T(t)\cdot Y(t)
\end{equation}
FORCE training procedure updates only the weight matrix \ineq{W} %(and to a certain extent \ineq{J}, see the discussion at the end of this section), 
such that the output is close to the target, i.e., \ineq{Z(t) = f(t)}.
The output error is defined as \ineq{e(t) = Z(t) - f(t)}.
%is fed back to the reservoir neurons.

A key characteristics of the FORCE training is that it keeps the error signal small by making rapid weight updates.
Let, weight updates are initiated at every \ineq{\Delta t} time interval.
Therefore, the weight update at time \ineq{t} is based on the output evaluated at its previous time step, i.e., at time \ineq{t-\Delta t}.
The output error at time \ineq{t} before performing the weight update is 
\begin{equation}
    \label{eq:error_pre}
    \footnotesize e_-(t) = Z(t-\Delta t) - f(t) = W^T(t-\Delta t)\cdot Y(t) - f(t)
\end{equation}
After the weight update, the output error is
\begin{equation}
    \label{eq:error_post}
    \footnotesize e_+(t) = W^T(t)\cdot Y(t) -f(t)
\end{equation}
Goals of the FORCE training are as follows.
\begin{enumerate}
    \item Reduce the output error during each weight update, i.e.,
    \begin{equation}
        \label{eq:error_reduction}
        \footnotesize |e_+(t)| < |e_-(t)|
    \end{equation}
    \item Converge the weight updates, i.e.,
    \begin{equation}
        \label{eq:convergence}
        \footnotesize \lim \frac{e_+(t)}{e_-(t)} = 1
    \end{equation}
\end{enumerate}
Using the recursive least square (RLS) algorithm for FORCE training~\cite{sussillo2009generating}, the weight update is given by
\begin{equation}
    \label{eq:weight_update}
    \footnotesize W^T(t) = W^T(t-\Delta t) - e_-(t)\cdot P(t)\cdot Y(t),
\end{equation}
where \ineq{P(t)} is updated during each weight update according to
\begin{equation}
    \label{eq:u_update}
    \footnotesize P(t) = P(t-\Delta t) - \frac{P(t-\Delta t) \cdot Y(t)\cdot Y^T(t)\cdot P(t-\Delta t)}{1+Y^T(t)\cdot P(t-\Delta t)\cdot Y(t)}.
\end{equation}
Here, \ineq{P(0) = \frac{{I}}{\alpha}}, where \ineq{{I}} is the identity matrix.

\noindent \textbf{Limitations:} {Although, FORCE can train the reservoir connections \ineq{J}, the learning is shown to be unstable due to the feedback. Additionally, FORCE training is shown to work only when the connectivity amongst the reservoir neurons is sparse.% and targeted for dynamic systems that can be modeled only using closed-form differential equations~\cite{force}.
}

\subsection{Training the Reservoir and Readout using full-FORCE}
Figure~\ref{fig:full_force} illustrates how full-FORCE algorithm is used to train a reservoir computing architecture~\cite{full_force}.
The original architecture is shown in Figure~\ref{fig:full_force}a. The dynamics of this architecture are as follows.
The output \ineq{Z(t)} is 
\begin{equation}
    \label{eq:output_ff}
    \footnotesize Z(t) = W^T\cdot Y(t),
\end{equation}
where \ineq{Y(t) = \mathcal{S}\big(X(t)\big)} is a non-linear function that implements sigmoidal neurons in the reservoir and 
\ineq{X(t)} is the neuron activation inside the reservoir.
%\ineq{W} is the weight matrix of the output layer. The neuron activation \ineq{X(t)} of the reservoir 
\ineq{X(t)} is governed by the first-order ordinary differential equation (ODE)
\begin{equation}
    \label{eq:activation_ff}
    \footnotesize \tau \frac{dX(t)}{dt} = -X(t) + J\cdot Y(t) + U_i\cdot f_{in}(t) + U_o\cdot Z(t),
\end{equation}
where \ineq{f_{in}(t)} is the input.
Observe that instead of using \ineq{f(t)} to represent both the input and target as in the case of the FORCE formulation (Sec.~\ref{sec:force}), we decouple the two.
This allows us to design a dynamical system that may receive a specific input (or no input at all) and generate a target that can be different from the input. This is fundamental to mimicking systems whose input-output relationship cannot be modeled using a closed-form differential equation.
%, and \ineq{U_o} is the output matrix connecting the output back to the reservoir using the feedback path.
% \begin{itemize}
% \color{red}{\item Since the feedback is a vector of weights sent to modify the reservoir weights, it is not a separate sub network in our architecture which requires training. If it is replaced with a trainable RSNN there is provision for increased computational cost and time and reduced efficacy if used as a feedback loop to train firing units of SNNs. } 
% \end{itemize}
The training objective of full-FORCE is to optimize \ineq{W} and \ineq{J} such that \ineq{Z(t) = f_{out}(t)}, where \ineq{f_{out}(t)} is the response of the dynamical system that we aim to model using the reservoir computing architecture.

The full-FORCE training procedure works as follows. 
A second reservoir network (called \underline{target-generating network}) is introduced during training as shown in Figure~\ref{fig:full_force}b.
This network is used to generate the target for every neuron in the original reservoir (henceforth referred to as \underline{task-performing network}).
The target-generating network is also a reservoir of randomly-connected recurrent neurons with fixed connections \ineq{J^D}. This reservoir is driven by both the input \ineq{f_{in}(t)} and the output \ineq{Z(t)}.
The rich dynamics of a randomly connected network 
%operating in the chaotic regime 
allows to model many dynamical systems by  suppressing chaos in a controlled manner using the two signals \ineq{f_{in}(t)} and \ineq{Z(t)}.% as shown in Figure~\ref{fig:full_force}b.

\begin{figure*}[h!]
	\centering
	%\vspace{-5pt}
	\centerline{\includegraphics[width=1.99\columnwidth]{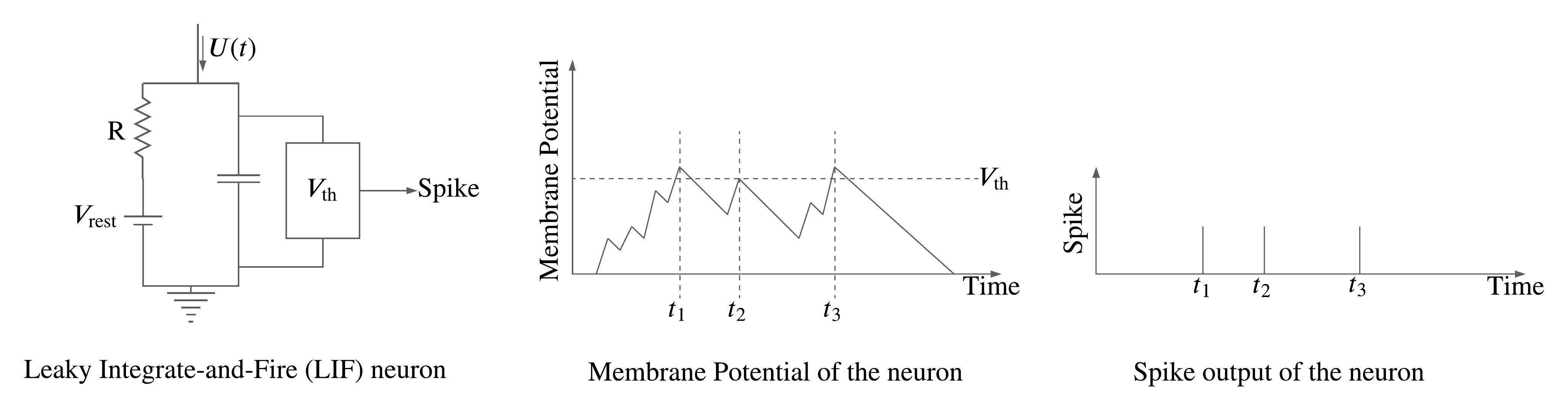}}
	%\vspace{-10pt}
	%\caption{An example of spiking neural network.}
	\caption{A leaky integrate-and-fire (LIF) neuron with current input $I(t)$ (left). The membrane potential over time of the neuron (middle). The spike output of the neuron representing its firing time (right).}
	%\vspace{-5pt}
	\label{fig:lif}
\end{figure*}

During training, dynamics of the task-performing and the target-generating networks are 
%given by
\begin{equation}
    \label{eq:reser1_ff}
    \footnotesize \tau \frac{dX(t)}{dt} = -X(t) + J\cdot Y(t) + U_i\cdot f_{in}(t)
\end{equation}

\begin{equation}
    \label{eq:reser2_ff}
    \footnotesize \tau \frac{dX^D(t)}{dt} = -X^D(t) + J^C\cdot Y^D(t) + U_i\cdot f_{in}(t) + U_o\cdot Z(t),
\end{equation}
where \ineq{X^D(t)} is the neuron activation in the target-generating network and \ineq{Y^D(t) = F\big(X^D(t)\big)} is the corresponding output of the sigmoidal neurons. The output response from the task-performing network is
\begin{equation}
    \label{eq:output_ff}
    \footnotesize Z(t) = W^T\cdot Y(t)
\end{equation}

Subsequently, FORCE training is used to learn both connections \ineq{W} and \ineq{J}. To train \ineq{W}, the target for \ineq{Z(t)} is \ineq{f_{out}(t)}, which is the response of the dynamical system that we aim to model. \ineq{W} is optimized using Equations~\ref{eq:weight_update} \& \ref{eq:u_update}.

To train \ineq{J}, we set \ineq{Y^D(t)} as the target for \ineq{Y(t)}. Equations~\ref{eq:weight_update} \& \ref{eq:u_update} are rewritten for the connectivity matrix \ineq{J} as 
\begin{equation}
    \label{eq:weight_update_ff}
    \footnotesize J(t) = J(t-\Delta t) - e_-^D(t)\cdot P^D(t)\cdot Y(t),
\end{equation}
where the error signal \ineq{e_-^D(t)} is 
\begin{equation}
    \label{eq:error_ff}
    \footnotesize e_-^D(t) = J(t-\Delta t)\cdot Y(t) - J^D(t)\cdot Y^D(t) - U_o\cdot Z(t),
\end{equation}
and the matrix \ineq{P^D(t)} is
\begin{equation}
    \label{eq:u_update_ff}
    \footnotesize P^D(t) = P^D(t-\Delta t) - \frac{P^D(t-\Delta t) \cdot Y(t)\cdot Y^T(t)\cdot P^D(t-\Delta t)}{1+Y^T(t)\cdot P^D(t-\Delta t)\cdot Y(t)}
\end{equation}

All formulations in this section use sigmoidal neurons, which are not efficient for neuromorphic computing.
%, which are typically not supported on event-driven neuromorphic hardware platforms. 
Next, we discuss how to implement full-FORCE for spiking neurons.

\section{Training RSNNs with full-FORCE}\label{sec:spiking_conversion}
In this section, we discuss how the full-FORCE training procedure is implemented for recurrent \underline{spiking} neural networks (RSNNs) with rate coding (Section~\ref{sec:rate_coding}) and time-to-first-spike (TTFS) coding (Section~\ref{sec:ttfs}).

\subsection{full-FORCE Training with Rate Coding}\label{sec:rate_coding}
The FORCE and full-FORCE formulation of Section~\ref{sec:prelim} uses the sigmoid neurons,
%These is utilized in
%i.e., Specifically, in 
%computing 
i.e., in computing \ineq{Y(t) = \mathcal{S}\big(x(t)\big)}, the function \ineq{\mathcal{S}(\cdot)} is a `S'-shaped function such as \texttt{tanh}, \texttt{ReLU} and \texttt{sigmoid}. Here, we formulate the full-FORCE training procedure for spiking neurons~\cite{maass1997networks}.

Spiking Neural Networks (SNNs) enable powerful computations due to their spatio-temporal information encoding capabilities~\cite{maass1997networks}.
An SNN consists of neurons, which are connected via synapses. A neuron can be implemented as an leaky integrate-and-fire (IF) logic, which is illustrated in Figure~\ref{fig:lif} (left).
Here, an input current \ineq{I(t)} (i.e., spike from a pre-synaptic neuron) raises the membrane voltage of the neuron. When this voltage crosses a threshold \ineq{V_{th}}, the IF logic emits an output spike, which propagates to is post-synaptic neuron. 
At the spike onset, the membrane capacitance discharges and the membrane voltage resets to the resting potential \ineq{V_{rest}}. This spike firing process repeats every time there is a current at the input.

Figure~\ref{fig:lif} (middle) illustrates the membrane voltage of the IF neuron due to input spike trains. Moments of threshold crossing are illustrated in Figure~\ref{fig:lif} (right). These are the firing times of output spike trains generated from the neuron.
An LIF spiking neuron is also associated with a refractory period, which is defined as the moment of silence (i.e., without any activity) following the firing of a spike.

\begin{figure*}[h!]
	\centering
	%\vspace{-5pt}
	\centerline{\includegraphics[width=1.99\columnwidth]{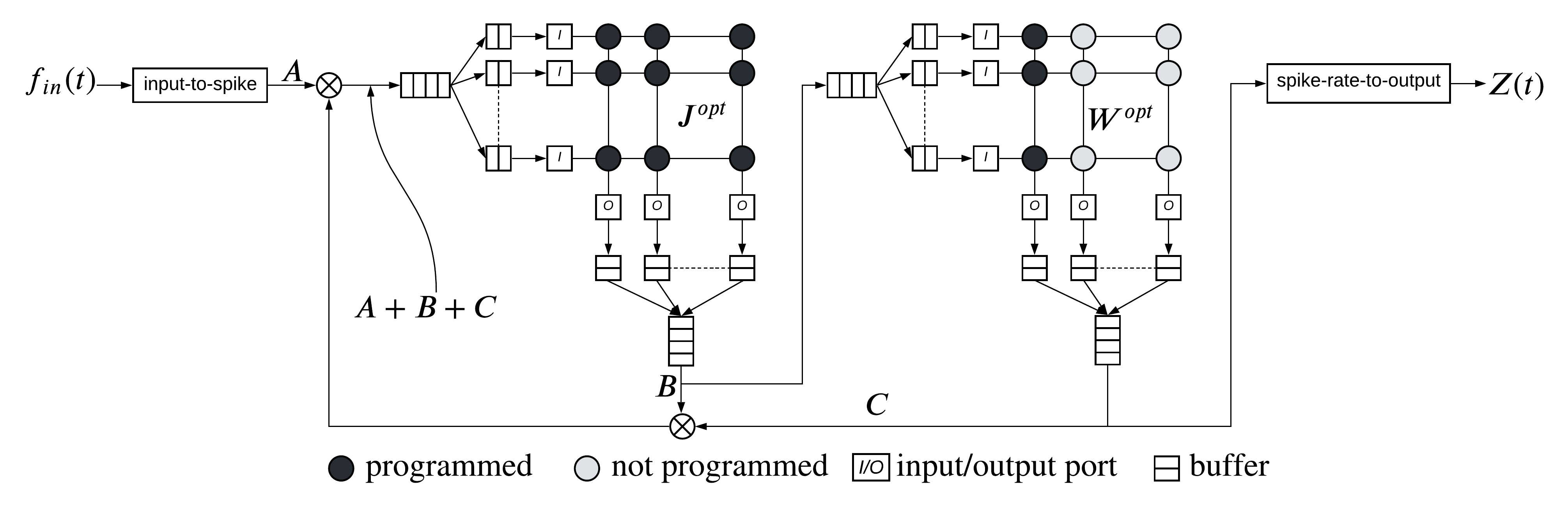}}
	%\vspace{-10pt}
	%\caption{An example of spiking neural network.}
	\caption{Implementation of the proposed trained RSNN on crossbar-based neuromorphic hardware.}
	%\vspace{-5pt}
	\label{fig:implementation}
\end{figure*}

The LIF neuron model provides a formalism of neuron firing rate in terms of a neuron's membrane time constant, the threshold voltage and the refractory period. Therefore, for full-FORCE training of LIF neurons, \ineq{Y(t)} in Equations~\ref{eq:weight_update}, \ref{eq:u_update}, \& \ref{eq:reser1_ff}-\ref{eq:u_update_ff} are all interpreted in terms of spike rate.

The LIF neuron model is described by the dynamics of a neuron's membrane voltage \ineq{V(t)} due to input current \ineq{I(t)} as
\begin{equation}
    \label{eq:neuron_membrane_voltage}
    \footnotesize C\frac{dV(t)}{dt} = I_\text{Bias} + I(t),
\end{equation}
where \ineq{C} is the membrane capacitance and \ineq{I_\text{Bias}} is the bias current. If \ineq{V(t) \ge V_{th}}, the neuron fires a spike, which  is represented as~\cite{meffin2004analytical}
\begin{equation}
    \label{eq:current_spike}
    \footnotesize I_\text{spike}(t) = C\left[\frac{dV(t)}{dt}\right]_{V(t) = V_{th}}^{-1}(V_\text{rest}-V_{th})\cdot\delta\Big(V(t)-V_{th}\Big)
\end{equation}
%Let \ineq{V(t) \ge V_{th}} be the membrane voltage of a neuron due to input current \ineq{I(t)}. 
Following a spike firing, the membrane voltage leaks over time. The voltage dynamics is 
\begin{equation}
    \label{eq:membrane_voltage_dynamics}
    \footnotesize V(t) = I(t)\cdot R\cdot \left[1-e^{\frac{-t}{\tau}}\right],
\end{equation}
where \ineq{R} is the membrane resistance and \ineq{\tau = \frac{1}{R\cdot C}} is the membrane time constant.
%, and \ineq{C} is the membrane capacitance.

The inter-spike interval, defined as the time it takes to reach the membrane potential \ineq{V_{th}} is calculated as
\begin{equation}
    \label{eq:isi}
    \footnotesize t_\text{isi} = -\tau \ln{\left(1-\frac{V_{th}}{I(t)\cdot R}\right)} = -\tau \ln{\left(1-\frac{I_\text{rh}}{I(t)}\right)},
\end{equation}
where \ineq{I_\text{rh}} is
the rheobasic current, defined as the smallest current that drives the membrane voltage to \ineq{V_{th}}, i.e.,
\begin{equation}
    \label{eq:rh}
    \footnotesize I_\text{rh} = \frac{V_{th}}{R}
\end{equation}

The spike firing rate is 
\begin{equation}
    \label{eq:spike_freq}
    \footnotesize R(t) = G\Big(I(t)\Big) =  \frac{1}{\tau_\text{ref} + t_\text{isi}} = \frac{1}{\tau_\text{ref} - \tau\ln{\left(1-\frac{I_\text{rh}}{I(t)}\right)}},
\end{equation}
where \ineq{\tau_\text{ref}} is the refractory period.

Considering current synapses, the input to a neuron is computed as
\begin{equation}
    \label{eq:cuba}
    \footnotesize I(t) = \sum_{i} w_i \sum_{k}\delta(t-t^i_k),
\end{equation}
where \ineq{w_i} is the synaptic weight connecting the \ineq{i^\text{th}} input neuron, and \ineq{\sum_{k}\delta\big(t-t^i_k\big)} are the spike trains of this input neuron with \ineq{t^i_k} being the time of its \ineq{k^\text{th}} spike. Here, \ineq{\delta(.)} is the Dirac Delta function used to represent a spike. 
Considering \ineq{\bigintsss \delta(t) = 1} and a time interval \ineq{\Delta t} for evaluating the spike rate, 
\begin{equation}
    \label{eq:cuba_final}
    \footnotesize I(t) = \sum_i w_i\cdot R_i\cdot \Delta t,
\end{equation}
where \ineq{R_i} is the spike rate of the \ineq{i^\text{th}} input neuron.% in the time interval \ineq{\Delta t}.

To explain the full-FORCE training procedure for LIF neurons, consider the input \ineq{f_{in}(t)} at time \ineq{t} be converted to spike trains using an analog-to-spike converter~\cite{HeartEstmNN}. It samples the instantaneous value of the input at time \ineq{t} and uses it to create spike trains whose inter-spike interval follows a Poisson process with a mean firing rate equal to the sampled value of the input. 
These Poisson spikes excite spiking neurons in the two reservoirs of Figure~\ref{fig:full_force}b.

Output of the task-performing network is
\begin{equation}
    \label{eq:out_ff_rsnn}
    \footnotesize Z(t) = G\Big(W^T\cdot R(t)\cdot \Delta t\Big),
\end{equation}
where \ineq{R(t)} is the spike rate of neurons in the task-performing network, \ineq{R(t)\cdot \Delta t} computes the total number of spikes, \ineq{W^T\cdot R(t)\cdot \Delta t} computes the current (see Eq.~\ref{eq:cuba_final}), and \ineq{G(\cdot)} computes the resultant spike rate using Equation~\ref{eq:spike_freq}.

Dynamics of the task-performing network is
\begin{equation}
    \label{eq:reser1_ff_rsnn}
    \footnotesize C\frac{dV(t)}{dt} = I(t) = \Big(J\cdot R(t) + U_{i}^T\cdot F_{in}(t)\Big)\cdot \Delta t,
\end{equation}
where \ineq{F_{in}(t) = f_{in}(t)\cdot \mathbf{I}} with \ineq{\mathbf{I}} being the identity matrix.
Observe the similarity of Equations~\ref{eq:reser1_ff_rsnn} and \ref{eq:reser1_ff}.

Next, the membrane voltage is evaluated and output spike trains are generated using Equation~\ref{eq:current_spike}. Finally, the output spike rate \ineq{R(t)} is computed using Equation~\ref{eq:spike_freq}.

Using a similar formulation, dynamics of the target-generating network is computed as
\begin{equation}
    \label{eq:reser2_ff_rsnn}
    \footnotesize C\frac{dV^D(t)}{dt} = I^D(t) = \Big(J^D\cdot R^D(t) + U_{i}^T\cdot F_{in}(t) + U_o\cdot Z(t)\Big)\cdot \Delta t,
\end{equation}

We apply full-FORCE weight updates as follows.
\begin{enumerate}
    \item \ineq{W} updates are performed using Equations~\ref{eq:weight_update} \& \ref{eq:u_update}.
    \item \ineq{J} updates are performed using Equations~\ref{eq:weight_update_ff} \& \ref{eq:u_update_ff}.
\end{enumerate}

In Section~\ref{sec:ttfs}, we show how the LIF neurons can be implemented in hardware and mapped to a platform using NeuroXplorer~\cite{neuroxplorer} or other similar frameworks.

% Let \ineq{F_i(t)} be a matrix defined as
% \begin{equation}
%     \label{eq:input_ff_snn}
%     \footnotesize F_i(t) = f_{in}(t)\cdot \mathbf{I}^{N\times N},
% \end{equation}
% where \ineq{\mathbf{I}^{N\times N}} is an \ineq{N\times N} identity matrix.

% Let \ineq{S\in \mathbb{R}^{N\times 1}} be the column vector representing the number of spikes generated by neurons of the reservoir at time \ineq{t}. Therefore, the total input current of the original reservoir neurons are
% \begin{equation}
%     \label{eq:currents_ff_reservoir}
%     \footnotesize I(t) = J\cdot S + U_i^T\cdot F_i \text{ (Equivalent of Equation~\ref{eq:reser1_ff})},
%  \end{equation}
%  where \ineq{J\in \mathbb{R}^{N\times N}} is the connectivity matrix of the original reservoir. The membrane voltage is then computed using Equation~\ref{eq:neuron_membrane_voltage}. Next, the membrane voltage is evaluated and output spike train is generated using Equation~\ref{eq:current_spike}.
 
%  The output spike current from the original reservoir are represented as 
%  \begin{equation}
%      \label{eq:out_currents_ff_reservoir}
%      \footnotesize I_\text{spike}(t+\Delta t) = 
%  \end{equation}

\subsection{Hardware Implementation}\label{sec:ttfs}
Consider the proposed full-FORCE training procedure is implemented in software using PyCARL~\cite{pycarl}, CARLsim~\cite{carlsim}, or Nengo~\cite{nengo}.
Subsequently, the trained weights (\ineq{J^{opt}} and \ineq{W^{opt}}) are mapped to the hardware.
Figure~\ref{fig:implementation} shows one possible implementation of a trained RSNN on a crossbar-based neuromorphic hardware such as the DYNAPs~\cite{dynapse}.
In the future, we will demonstrate implementation on other architectures such as Loihi~\cite{loihi}, and $\mu$Brain~\cite{sentryos}.

% \begin{figure}[h!]
% 	\centering
% 	%\vspace{-5pt}
% 	\centerline{\includegraphics[width=0.99\columnwidth]{images/implementation_v2.pdf}}
% 	%\vspace{-10pt}
% 	%\caption{An example of spiking neural network.}
% 	\caption{Implementation of the proposed trained RSNN on crossbar-based neuromorphic hardware.}
% 	%\vspace{-5pt}
% 	\label{fig:implementation}
% \end{figure}

A crossbar is a two dimensional organization of wordlines and bitlines.
A synaptic cell is placed at the intersection of each wordline and bitline~\cite{hu2014memristor}.
%where a synaptic weight is programmed. 
A synaptic cell can be designed using non-volatile memory (NVM)~\cite{mallik2017design}.

Buffers are implemented at the input and output of a crossbar to store spikes, which are encoded in Address Event Representation (AER) format.
Each input port of a crossbar (marked 'I' in the figure) generates a voltage pulse for each AER packet stored in its input buffer. 
This voltage is multiplied with the conductance of a synaptic cell to generate a current. 
On an output port, which represents a post-synaptic neuron, current from different input ports are integrated (Kirchoff's current law). This is the input current \ineq{I(t)} which modulates a neuron's membrane voltage \ineq{V(t)} using the LIF dynamics described in Equations~\ref{eq:neuron_membrane_voltage}-\ref{eq:spike_freq}.
Each output spike is converted to AER format and stored in the  output buffer for transmission over the shared communication channel.

The RSNN architecture can be implemented on two \ineq{N\times N} crossbars as shown in the figure. 
Here \ineq{N} is the number of neurons in the reservoir of Figure~\ref{fig:full_force}.
These crossbars implement matrices \ineq{J^{opt}} and \ineq{W^{opt}}, respectively.
%For each crossbar, we show the synaptic weights that are programmed.

The reservoir network (crossbar 1) receives input spikes from three sources -- the input signal \ineq{f_{in}(t)} converted to spike (marked \ineq{A}), the recurrent connection from other neurons in the reservoir (marked \ineq{B}), and the output from the readout (crossbar 2) fed back to the reservoir (marked \ineq{C}).
Observe, the recurrent connections of the reservoir neurons are implemented as a feedforward architecture in crossbar 1 with the output of the crossbar fed back to the input.
The spike rate computed at the output of crossbar 2 is interpreted as the output \ineq{Z(t)} of the RSNN architecture.

%\subsubsection{\underline{Implementation Specifics}}
The design of Figure~\ref{fig:implementation} can be implemented on a custom hardware using the Neural Engineering Framework (NEF)~\cite{nef}, NeuroXplorer~\cite{neuroxplorer}, and SpiNeMap~\cite{spinemap}.
Alternatively, the SODASNN framework can be used to synthesize the above design on a Field Programmable Gate Array (FPGA) device~\cite{shihao_soda}.
We limit our discussion to implementing an RSNN on a crossbar-based neuromorphic hardware. To this end, the design flow of~\cite{shihao_designflow,dfsynthesizer_pp,dfsynthesizer} can be used. A comprehensive overview of other frameworks and mapping flows can be found in~\cite{huynh2022implementing}.
We use NeuroXplorer~\cite{neuroxplorer}.
For crossbar-based designs, spike communication between the computational units (input unit, crossbars, and output unit) takes place via a shared interconnect such as Network-on-Chip (NoC) or Segmented Bus~\cite{neunoc,balaji2019exploration,catthoor2018very}. 
%Therefore, \ineq{\bigotimes} in the figure represents time-multiplexing on the interconnect.

A critical limitation of a crossbar-based neuromorphic hardware is the constraint on the size of a crossbar. Crossbars are designed as an \ineq{N\times N} hardware with \ineq{N} inputs and \ineq{N} outputs. The typical value of \ineq{N} ranges between 128 and 256~\cite{ankita_igsc,song2022design,twisha_energy}.
Therefore, if a reservoir contains more neurons than what can be accommodated on a crossbar, the reservoir network needs to be decomposed into smaller sub-networks using for instance, the spatial decomposition technique of~\cite{esl20}.
Subsequently, each decomposed network can be mapped to an individual crossbar of the hardware.
To demonstrate our full-FORCE learning, here we use crossbar sizes that can map each reservoir in its entirety, i.e., no spatial decomposition is necessary. 
Area-accuracy tradeoffs associated with spatial decomposition for full-FORCE will be addressed in the future.

Although using LIF neurons with rate coding simplifies the implementation, such coding generally leads to a higher number of spikes~\cite{park2020t2fsnn} (see our results in Section~\ref{sec:results}).
Following are the two critical neuromorphic design issues associated with having more spikes. 
First, higher spike count leads to an increase in the network congestion (i.e., delay) and energy consumption on the shared interconnect~\cite{barchi2018mapping,urgese2016optimizing,psopart}.
Second, it increases the buffer requirement on input and output ports of each crossbar~\cite{shihao_designflow}, which increases the design cost (area and power) of an RSNN implementation.

\subsection{full-FORCE Training with TTFS Coding}

In this coding scheme, the first spike generated from a neuron is prioritized over subsequent spikes from the same neuron~\cite{park2020t2fsnn}.
Implementation-wise, a neuron follows the LIF dynamics described in Section~\ref{sec:rate_coding}. However, there are two modifications necessary to implement TTFS coding.

First, instead of keeping the threshold voltage constant as in rate coding, here the threshold voltage \ineq{V_{th}} is time dependant and is described using an exponential function
\begin{equation}
\label{eq:threshold_voltage_ttfs}
\footnotesize
%V_{th}(t) =\theta_0 \exp(−t/\tau),  
V_{th}(t) = \theta_0 \cdot e^{\frac{-t}{\tau}},
\end{equation}
where \ineq{\theta_0} is a constant.

Second, inhibition behavior is implemented such that once a neuron fires a spike, it is inhibited from generating spikes until the next input arrives~\cite{goltz2019fast}.
To do so, we introduce spike weights \ineq{w_s(t)} in Equation~\ref{eq:cuba}.
Specifically, the modified spike trains from a post-synaptic neuron are represented as 
\begin{equation}
    \footnotesize \text{spike trains} = w_s(t)\sum_{k}\delta(t-t^i_k),
\end{equation}
where
\begin{equation}
\label{eq:ttfsspikeweight}
\footnotesize
w_s(t) = e^{\frac{-t}{\tau_s}}.
\end{equation}
Here, \ineq{\tau_s} is the time constant for spike inhibition. % of the spike weights. %This is to inhibit spikes from the post-synaptic neuron after generating the fist spike.
% carries the most weight, and transmits greater information to post-synaptic neurons. The post-synaptic neurons accumulate their membrane potentials by integrating the synaptic input and output spikes when the potentials $>$  firing threshold~\cite{guo2021neural}.
% The input is  normalized and   threshold $V_{th}$ is determined by exponential function
% \begin{equation}
% \label{eq:thresholdvoltage}
% \footnotesize
% %V_{th}(t) =\theta_0 \exp(−t/\tau),  
% V_{th}(t) = \theta_0 \cdot e^{\frac{-t}{\tau}}
% \end{equation}
% where \ineq{\theta_0} is threshold constant =1. Upon exceeding the threshold,a spike is generated and the input is inhibited from generating further spikes. 

% Considering the  time of the first spike, the dynamics of the post-synaptic potential is
% %generated is tracked to generate the sum of post-synaptic potentials by 
% \begin{equation}
% \label{eq:ttfssynapse}
% \footnotesize
% z_j(t)=\sum_{i}V_{ij}(t)=w_s(t)\sum_{i}w_{ij}s_i(t),
% \end{equation}
% where \ineq{z_j(t)} is the synaptic input to the postsynaptic neuron \ineq{j}, \ineq{V_{ij}(t)} is the post-synaptic potential resulting from the input neuron i, \ineq{s_i(t)} is the input spike train from the presynaptic neuron i, \ineq{w_{ij}(t)} is the synaptic weight. \ineq{w_s(t)} is the spike weight at time t and is given by 
% \begin{equation}
% \label{eq:ttfsspikeweight}
% \footnotesize
% w_s(t) = e^{\frac{-t}{\tau_s}}, 
% \end{equation}
% At each time step, only the first spike is forwarded while others are suppressed. This reduces the number of spikes and the time it takes for a signal to converge. 

\section{Evaluation Framework}\label{sec:evaluation}

\subsection{Algorithm Implementation}
We implemented full-FORCE training in Python and simulated it on a Lambda workstation, which has AMD Threadripper 3960X with 24 cores, 128 MB cache, 128 GB RAM, and two RTX3090 GPUs. The code is available online at \url{https://github.com/drexel-DISCO/SNN-full-FORCE}. 
% \begin{itemize}
% \color{blue}{\item We implemented the full-FORCE training procedure using the CARLsim~\cite{carlsim} simulator, facilitates parallel simulation of large SNNs using CPUs and multi-GPUs.
% The framework is simulated on a Lambda workstation, which has AMD Threadripper 3960X with 24 cores, 128 MB cache, 128 GB RAM, and 2 RTX3090 GPUs.
% Using the PyCARL interface of CARLsim~\cite{pycarl}, the implementation can be easily integrated with PyNN~\cite{pynn} and simulate in other spiking simulators such as Brian~\cite{brian} and NEURON~\cite{neuron}.}
%Upon acceptance, we will release the GitHub repository of the full-FORCE implementation in CARLsim to foster future research.%}
%\end{itemize}
\subsection{Evaluated Dynamical Systems}
We evaluate 8 dynamical system which are as follows.
\begin{enumerate}
    \item \texttt{Sine:} The sine function \ineq{y(t) = \texttt{sine}(\omega\cdot t)} is commonly used to represent a periodic dynamical system.
    \item \texttt{Sum of Sines:} This is the sum of two sine functions with 4Hz and 6Hz frequencies, respectively.
    \item \texttt{Product of Sines:} This is the product of two sine functions with 4Hz and 6Hz frequencies, respectively.
    \item \texttt{Accordian:} This is an oscillatory signal that repeats periodically every 2s. This is represented using \ineq{y(t) = \texttt{sine}(\omega\cdot t)}. However, the frequency increases linearly from \ineq{2\pi} to \ineq{6\pi} Hz for the first half of the oscillation period, and then decreases linearly from \ineq{6\pi} to \ineq{2\pi} Hz during the second half of the oscillation period. 
    \item \texttt{Ode-to-Joy:} This is the 5-component signal representing the Ode-to-Joy notes by Beethoven, where each component corresponds to musical notes \texttt{C-G}. A note is represented with an upward pulse in the signal. Quarter notes are the positive parts of a 2 Hz sine wave and half notes are the positive parts of a 1 Hz sine wave. 
    \item \texttt{Triangle:} This is a triangular-shaped waveform that is represented by \ineq{y(t) = \texttt{(2a/$\pi$){arcsin}({sine}(2$\pi$}/p)x)}.
    \item \texttt{Van der Pol Harmonic:} This is a non-con\-servatory oscillator discovered by Van der Pol and is represented as \ineq{\ddot x=\mu(1-x^2)\dot x -x} with \ineq{\mu=0.3}. 
    \item \texttt{Van der Pol Relaxed:} This is a relaxed form of the Van der Pol oscillator with \ineq{\mu} set to 0.5. 

\end{enumerate}

\begin{figure}[h!]
	\centering
	%\vspace{-10pt}
	\centerline{\includegraphics[width=1.09\columnwidth]{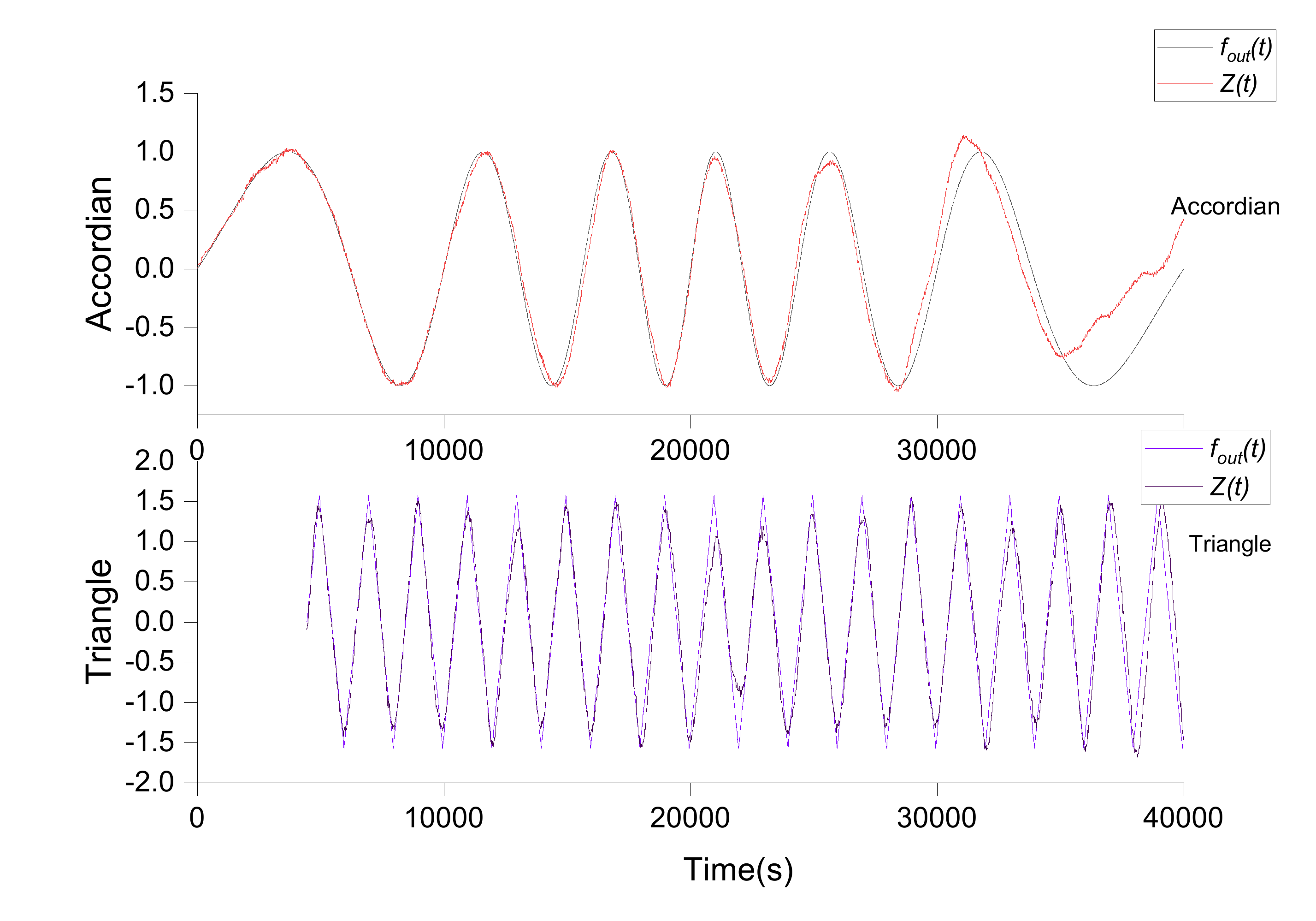}}
	%\vspace{-10pt}
	%\caption{An example of spiking neural network.}
	\caption{Target $f_{out}$(t) and response $Z(t)$ for the \texttt{Accordian} function (top) and \texttt{Triangule} function (bottom) using full-FORCE with rate coding.}
	\label{fig:at}
\end{figure}

Figures~\ref{fig:at} (top and bottom) shows the target \ineq{f_{out}(t)} and response \ineq{Z(t)} for the dynamical systems representing the \texttt{Accordian} and \texttt{Triangle} functions, respectively.
%, while Figure~\ref{fig:otj} shows the same for the dynamical system representing the five components of \texttt{Ode-to-Joy}.

% \begin{figure}[h!]
% 	\centering
% 	%\vspace{-10pt}
% 	\centerline{\includegraphics[width=0.99\columnwidth]{images/otj.pdf}}
% 	%\vspace{-10pt}
% 	%\caption{An example of spiking neural network.}
% 	\caption{Target $f_{out}$(t) and response $Z(t)$ for the five component \texttt{Ode-to-Joy} function on the top and \texttt{Triangule} function on the bottom using full-FORSE with rate coding.}
% 	\label{fig:otj}
% \end{figure}

%Figure~\ref{fig:signals} shows these dynamic systems.

\subsection{Evaluated Training Procedure}
We evaluate the following three training procedures.
\begin{enumerate}
    \item \texttt{FORCE-LIF-Rate:} This is the FORCE training with LIF neurons and rate coding~\cite{force_snn}. Here, only the readout weights \ineq{W} are trained. 
    %We overcome the limitations discussed in Section~\ref{sec:limitations_sota}.
    \item \texttt{full-FORCE-LIF-Rate:} This is our full-FORCE training with LIF neurons and rate coding. Here, both the reservoir wights \ineq{J} and output weights \ineq{W} are trained.
    \item \texttt{full-FORCE-LIF-TTFS:} This is our full-FORCE training with LIF neurons and TTFS coding.
\end{enumerate}

\subsection{Evaluated Metrics} We evaluate the following three metrics
\begin{enumerate}
    \item \texttt{Mean Square Error (MSE):} This is defined as the sum of the squared difference of the target and prediction, averaged over a fixed number of time steps. This is computed as follows.
    \begin{equation}
        \label{eq:mse}
        \footnotesize \texttt{MSE} = \frac{1}{n} \sum_{i=1}^n \Big(Z(t) - f_{out}(t) \Big)^2
    \end{equation}
    %\item \texttt{Normalized Error (NE):}
    
    \item \texttt{Time to Converge (TTC):} This is the number of epochs it takes to train an RSNN to mimick a target dynamical system.
    
    \item \texttt{Average Spike Rate (SR):} This is the total number of spikes generated by all neurons in an RSNN averaged over a time interval.
\end{enumerate}

\section{Results and Discussion}\label{sec:results}
\subsection{Mean Square Error (MSE)}\label{sec:mse_results}
Table~\ref{tab:mse} reports the MSE for the evaluated training procedures for all eight dynamical systems (lower MSE is better). The experimental setup is as follows. The number of neurons in the RSNN reservoir is set to 1000 for \texttt{FORCE-LIF-Rate} and \texttt{full-FORCE-LIF-Rate}. For \texttt{full-FORCE-LIF-TTFS}, this is set to 200. These numbers are decided based on the following considerations. The number of neurons for the full-FORCE training procedure (both rate coding and TTFS coding) is configured to keep the MSE less than 0.35 for all evaluated dynamical systems. This MSE constraint of 0.35 is user configurable. To achieve this constraint, the full-FORCE training procedure with rate coding requires 1000 neurons while TTFS coding requires only 200 neurons. We also set the number of neurons in the reservoir for the FORCE training with rate coding to 1000 (i.e., the same as full-FORCE with rate coding).
For all three training procedures, the number of training epochs is set to 50. We make the following three key observations.

\begin{table}[h!]
	\renewcommand{\arraystretch}{1.2}
	\setlength{\tabcolsep}{0.8pt}
	\caption{MSE of Proposed full-FORCE (rate and TTFS) against state-of-the-art FORCE.}
	\label{tab:mse}
	\vspace{-5pt}
	\centering
	\begin{threeparttable}
	{\fontsize{8}{10}\selectfont
	    %\vspace{-10pt}
		\begin{tabular}{c u{1.2cm} u{2.2cm} u{2.2cm}}
		    %\tabucline[2pt]{-}
			%\hline
			\Xhline{4\arrayrulewidth}
			%\hline
			& \textbf{FORCE-LIF-Rate (SOTA)} & \textbf{full-FORCE-LIF-Rate (Proposed)} & \textbf{full-FORCE-LIF-TTFS (Proposed)}\\
			\Xhline{4\arrayrulewidth}
			\texttt{Sine} & 0.041 & 0.009 & 0.32 \\
			\rowcolor{gainsboro} \texttt{Sum of Sines} & 0.470 & 0.250 & 0.240 \\
			\texttt{Product of Sines} & 0.393 & 0.330 & 0.330 \\
			\rowcolor{gainsboro} \texttt{Accordian} & 0.909 & 0.03 & 0.2\\
			\texttt{Ode-to-Joy} & 0.701 & 0.002 & 0.02\\
			\rowcolor{gainsboro} \texttt{Triangle} & 0.044 & 0.018 & 0.244 \\
			\texttt{Van der Pol H.} & 0.048 & 0.048 & 0.15 \\
			\rowcolor{gainsboro} \texttt{Van der Pol R.} & 0.009 & 0.008 & 0.08 \\
			%\tabucline[2pt]{-}
% 			\cite{lsm} & $\times$ & Readout Only (supervised) & $\times$ & $\surd$ & $\surd$\\
% 			\rowcolor{gainsboro} \cite{zhang2019spike} & $\times$ & Reservoir and Readout (BPTT) & $\times$ & $\surd$ & $\times$\\
% 			\cite{sussillo2009generating} & $\surd$ & Readout (FORCE) & $\surd$ & $\times$ & $\times$\\
% 			\rowcolor{gainsboro} \cite{full_force} & $\surd$ & Reservoir and Readout (full-FORCE) & $\surd$ & $\times$ & $\times$\\
% 			\cite{force_snn} & $\surd$ & Readout (FORCE) & $\surd$ & $\surd$ & $\times$\\
% 			\hline
% 			%\tabucline[2pt]{-}
% 			\rowcolor{azuremist} \textbf{ours} & $\surd$ & Reservoir and Readout (full-FORCE) & $\surd$ & $\surd$ & $\surd$\\
			%\hline
			\Xhline{4\arrayrulewidth}
			%\tabucline[2pt]{-}
	\end{tabular}}
	\end{threeparttable}
	%\vspace{12pt}
	\vspace{-5pt}
\end{table}

First, the MSE using FORCE is higher than the MSE constraint of 0.35 for four of the eight dynamical systems (\texttt{Sum of Sines}, \texttt{Product of Sines}, \texttt{Accordian}, and \texttt{Ode-to-Joy}). This means that for these systems a bigger reservoir (more than 1000 neurons) and possibly more training epochs are needed. For all eight dynamical systems, full-FORCE with rate coding satisfies the MSE constraint using 1000 neurons.

Second, the MSE using full-FORCE is lower than FORCE by an average of 51\% (between 0.1\% and 99.7\%). 
%This means that full-FORCE can better model dynamical systems than FORCE. 
This is due to its supervised training of both the reservoir weights \ineq{J} and readout weights \ineq{W}. 
The FORCE training, on the other hand, trains only the readout weights.
%can be used to efficiently train only the the output weights \ineq{W}
%in RSNNs when \ineq{J} is a fully-connected (i.e., not a sparse) matrix. 
%neurons in the reservoir are fully connected. FORCE training procedure has been shown to train the reservoir weights \ineq{J} when there is a sparse connectivity between neurons in the reservoir. 
Furthermore, neurons are fully-connected in the reservoir to model systems that cannot be represented using closed-form differential equations.
Therefore, the full-FORCE training procedure can train the reservoir more efficiently than FORCE.

Third, the MSE of full-FORCE with TTFS coding is higher than rate coding by a factor of 10x (on average).
Although the MSE is higher (meaning the performance is lower), full-FORCE with TTFS coding satisfies the MSE constraint of 0.35. The key advantage of the TTFS coding is the reduced number of neurons required to achieve the MSE constraint. With only 200 neurons, i.e., 5x lower number of neurons than rate coding, full-FORCE is able to meet the MSE constraint for all dynamical systems. This is an important consideration for the hardware implementation because fewer the number of neurons, smaller is the hardware area and power consumption.

\begin{figure}[h!]%
    \centering
    %\vspace{-10pt}
    \subfloat[$f_{out}(t)$ abd $Z(t)$ for full-FORCE training with rate coding.\label{fig:sos_1}]{{\includegraphics[width=9.2cm]{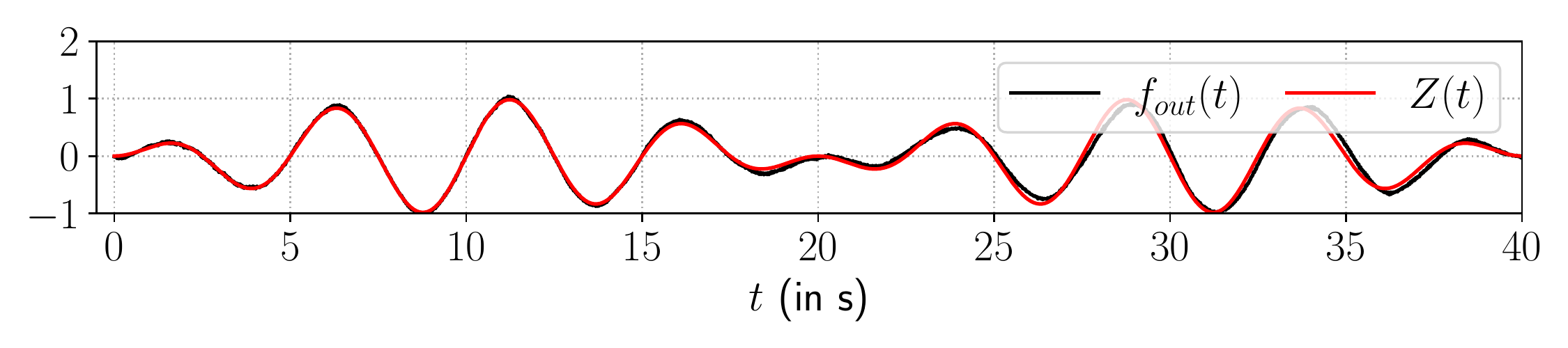} }}%
    \hfill
    \subfloat[$f_{out}(t)$ abd $Z(t)$ for full-FORCE training with TTFS coding.\label{fig:sos_2}]{{\includegraphics[width=9.2cm]{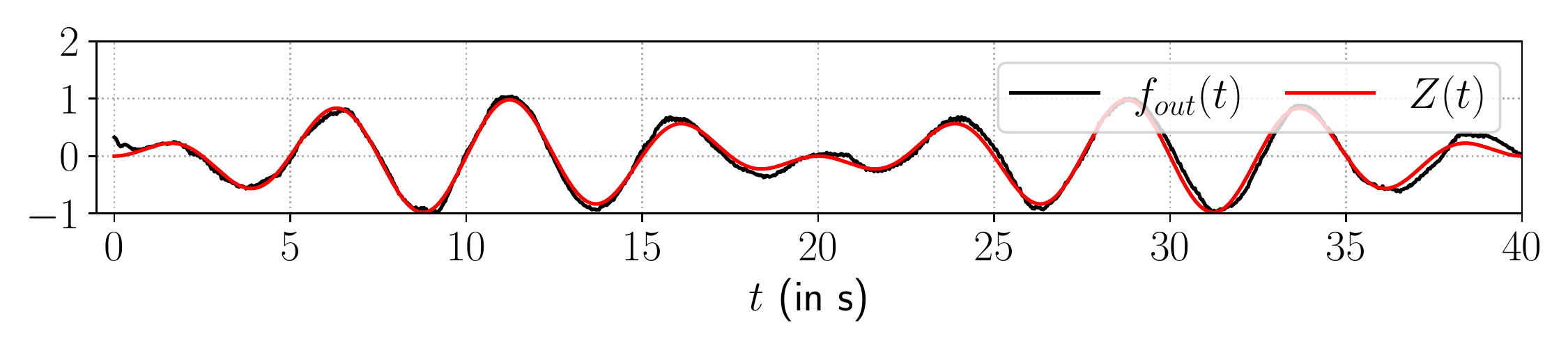} }}%
    %\subfloat[Scheduling sub-networks to \mubrain{} pipelines.\label{fig:lenet_mubrain_mapping}]{{\includegraphics[width=4.2cm]{images/lenet_platform_mapping.png} }}%
    %\vspace{-10pt}
    \caption{Comparative performance of rate coding and TTFS coding for the \texttt{sum-of-sine} dynamical system.}%
    \label{fig:compare_sos}%
    %\vspace{-10pt}
\end{figure}

To give more insights to these results, Figures~\ref{fig:compare_sos} shows the comparative performance of the full-FORCE training with rate coding and TTFS coding for the dynamical system represented using the \texttt{Sum of Sines} function.

Key takeaway points from these results are the following.
\begin{enumerate}
    \item With the same number of neurons in an RSNN reservoir (hardware constraint), the full-FORCE training with rate coding leads to a lower MSE than FORCE.
    \item To achieve a given MSE constraint, the full-FORCE with TTFS coding requires fewer neurons than both the FORCE and full-FORCE with rate coding.
\end{enumerate}

\subsection{Time to Converge}\label{sec:ttc}
Table ~\ref{tab:ttc} reports the number of epochs it takes  for each training procedure to reach an MSE of 0.25 in each epoch with 200 neurons. We make three key observations. First, 
between FORCE and full-FORCE with rate coding, FORCE requires a smaller number of epochs to converge.
%for FORCE trained networks the mean timestep to reach MSE $<$ 0.25 is lower than full-FORCE trained networks. 
Second, full-FORCE with TTFS coding requires fewer epochs (on average) to converge compared to full-FORCE with rate coding. 
Third, full-FORCE with TTFS coding requires more epochs to converge for \texttt{Van der Pol Harmonics} and \texttt{Van der Pol Relaxed} due to their deterministic chaotic behavior.
%There is a 40\% reduction in time for TTFS with respect to rate based full-FORCE.But for Van der Pol oscillations the full-FORCE TTFS takes larger time than FORCE, and is almost similar to ode to joy FORCE time, which can be a liability of their multi-components nature and deterministic chaotic behavior.

\begin{table}[h!]
	\renewcommand{\arraystretch}{1.2}
	\setlength{\tabcolsep}{0.8pt}
	\caption{Time to Converge of Proposed full-FORCE RSNN (rate and TTFS) against FORCE.}
	\label{tab:ttc}
	%\vspace{-5pt}
	\centering
	\begin{threeparttable}
	{\fontsize{8}{10}\selectfont
	    %\vspace{-10pt}
		\begin{tabular}{c u{1.2cm} u{2.2cm} u{2.2cm}}
		    %\tabucline[2pt]{-}
			%\hline
			\Xhline{4\arrayrulewidth}
			%\hline
			& \textbf{FORCE-LIF-Rate (SOTA)} & \textbf{full-FORCE-LIF-Rate (Proposed)} & \textbf{full-FORCE-LIF-TTFS (Proposed)}\\
			\Xhline{4\arrayrulewidth}
			\texttt{Sine} & 2.29 & 2.85 & 0.18 \\
			\rowcolor{gainsboro} \texttt{Sum of Sines} & 2.39 & 1.8 & 1.43 \\
			\texttt{Product of Sines} & 2.38 & 3.19 & 1.19 \\
			\rowcolor{gainsboro} \texttt{Accordian} & 1.55 & 1.95 & 1.39\\
			\texttt{Ode-to-Joy} & 3.06 & 4.48 & 3.05\\
			\rowcolor{gainsboro} \texttt{Triangle} & 2.33 & 2.86 & 1.49 \\
			\texttt{Van der Pol H.} & 3.52 & 4.9 & 4.15 \\
			\rowcolor{gainsboro} \texttt{Van der Pol R.} & 3.67 & 5.8 & 3.99 \\
	%\tabucline[2pt]{-}
% 			\cite{lsm} & $\times$ & Readout Only (supervised) & $\times$ & $\surd$ & $\surd$\\
% 			\rowcolor{gainsboro} \cite{zhang2019spike} & $\times$ & Reservoir and Readout (BPTT) & $\times$ & $\surd$ & $\times$\\
% 			\cite{sussillo2009generating} & $\surd$ & Readout (FORCE) & $\surd$ & $\times$ & $\times$\\
% 			\rowcolor{gainsboro} \cite{full_force} & $\surd$ & Reservoir and Readout (full-FORCE) & $\surd$ & $\times$ & $\times$\\
% 			\cite{force_snn} & $\surd$ & Readout (FORCE) & $\surd$ & $\surd$ & $\times$\\
% 			\hline
% 			%\tabucline[2pt]{-}
% 			\rowcolor{azuremist} \textbf{ours} & $\surd$ & Reservoir and Readout (full-FORCE) & $\surd$ & $\surd$ & $\surd$\\
			%\hline
			\Xhline{4\arrayrulewidth}
			%\tabucline[2pt]{-}
	\end{tabular}}
	\end{threeparttable}
	%\vspace{12pt}
	%\vspace{-10pt}
\end{table}

\subsection{Average Spike Rate}\label{sec:sr}
\begin{figure}[h!]
	\centering
	%\vspace{-10pt}
	\centerline{\includegraphics[width=0.99\columnwidth]{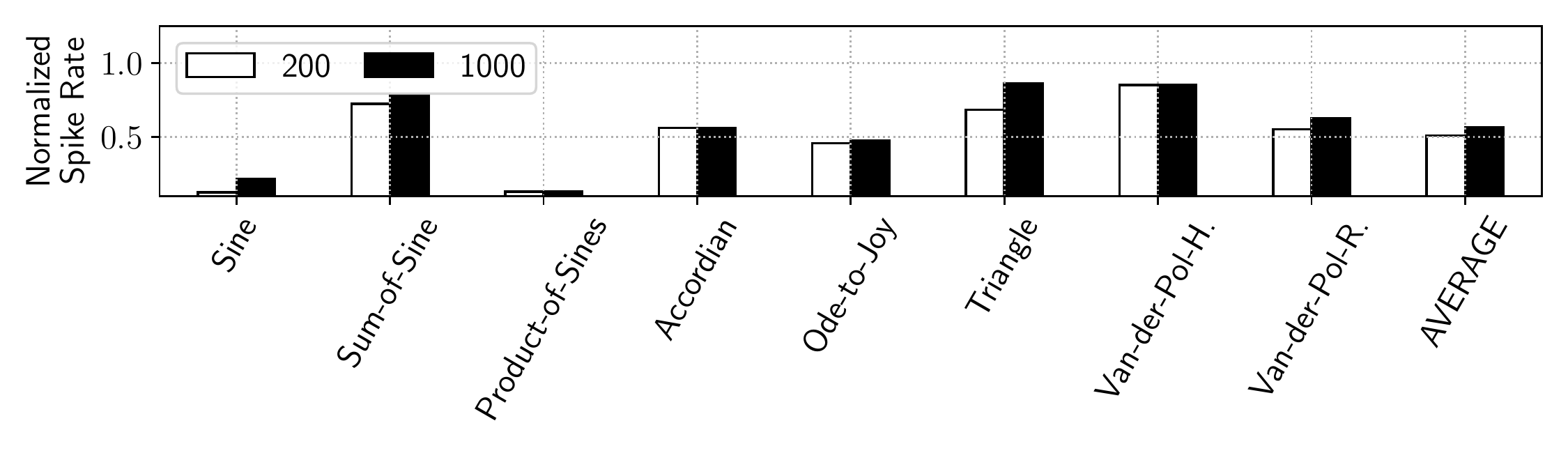}}
	%\vspace{-10pt}
	%\caption{An example of spiking neural network.}
	\caption{Spike rate for the evaluated dynamical systems with 200 and 1000 neurons in the RSNN reservoir.}
	\label{fig:sr}
\end{figure}

Figure~\ref{fig:sr} plots the spike rate of full-FORCE with TTFS coding normalized to rate coding for the eight evaluated dynamical systems. Results are reported for 200 and 1000 neurons in the RSNN reservoir. We make the following two key observations.

%\vspace{-10pt}

First, TTFS coding results in fewer spikes than rate coding for all two sizes of the reservoir and across all eight dynamical systems. This is because of the nature of TTFS coding, where instead of communicating all spikes as in rate coding, 
information is represented by the relative time of arrival of the spikes with respect to the first spike.
%In TTFS coding, neurons spike at most once generating fewer spikes.
On average, TTFS generates 50\% fewer spikes than rate coding.

Second, with more neurons in the reservoir, the number of spikes increases only marginally (average 15\%) for the TTFS coding.
The spike count for rate coding increases proportionately with the number of neurons.

\subsection{Noise Performance}\label{sec:noise}
We test the noise robustness using an RSNN reservoir with 600 neurons. We use 50 training epochs and add 10\% Gaussian noise to the input signal. 
%to determine the noise-robustness for the different training methods, the MSE of each are shown in 
Table ~\ref{tab:msenoise} reports the corresponding MSE.
We observe that full-FORCE (both rate and TTFS coding) has lower MSE than FORCE demonstrating its noise robustness. However, due to using a single spike to encode information, full-FORCE with TTFS coding has a higher MSE than rate coding.

\begin{table}[h!]
	\renewcommand{\arraystretch}{1.2}
	\setlength{\tabcolsep}{0.8pt}
	\caption{MSE of evaluated training procedures with Gaussian Noise of Proposed full-FORCE RSNN (rate based and TTFS) against  FORCE SNN (SOTA).}
	\label{tab:msenoise}
	\vspace{-5pt}
	\centering
	\begin{threeparttable}
	{\fontsize{8}{10}\selectfont
	    %\vspace{-10pt}
		\begin{tabular}{c u{1.2cm} u{2.2cm} u{2.2cm}}
		    %\tabucline[2pt]{-}
			%\hline
			\Xhline{4\arrayrulewidth}
			%\hline
			& \textbf{FORCE-LIF-Rate (SOTA)} & \textbf{full-FORCE-LIF-Rate (Proposed)} & \textbf{full-FORCE-LIF-TTFS (Proposed)}\\
			\Xhline{4\arrayrulewidth}
			\texttt{Sine} & 0.23 & 0.048 & 0.2 \\
			\rowcolor{gainsboro} \texttt{Sum of Sines} & 0.85 & 0.02 & 0.23 \\
			\texttt{Product of Sines} & 0.33 & 0.24 & 0.35 \\
			\rowcolor{gainsboro} \texttt{Accordian} & 1.75 & 1.72 & 1.739\\
			\texttt{Ode-to-Joy} & 0.75 & 0.05 & 0.135\\
			\rowcolor{gainsboro} \texttt{Triangle} & 0.08 & 0.074 & 0.25 \\
			\texttt{Van der Pol H.} & 0.223 & 0.17 & 0.199 \\
			\rowcolor{gainsboro} \texttt{Van der Pol R.} & 0.197 & 0.102 & 0.15 \\
	%\tabucline[2pt]{-}
% 			\cite{lsm} & $\times$ & Readout Only (supervised) & $\times$ & $\surd$ & $\surd$\\
% 			\rowcolor{gainsboro} \cite{zhang2019spike} & $\times$ & Reservoir and Readout (BPTT) & $\times$ & $\surd$ & $\times$\\
% 			\cite{sussillo2009generating} & $\surd$ & Readout (FORCE) & $\surd$ & $\times$ & $\times$\\
% 			\rowcolor{gainsboro} \cite{full_force} & $\surd$ & Reservoir and Readout (full-FORCE) & $\surd$ & $\times$ & $\times$\\
% 			\cite{force_snn} & $\surd$ & Readout (FORCE) & $\surd$ & $\surd$ & $\times$\\
% 			\hline
% 			%\tabucline[2pt]{-}
% 			\rowcolor{azuremist} \textbf{ours} & $\surd$ & Reservoir and Readout (full-FORCE) & $\surd$ & $\surd$ & $\surd$\\
			%\hline
			\Xhline{4\arrayrulewidth}
			%\tabucline[2pt]{-}
	\end{tabular}}
	\end{threeparttable}
	%\vspace{12pt}
	\vspace{-10pt}
\end{table}

\subsection{Interval Matching Task}\label{sec:extra_tasks}
We train an RSNN for an interval matching task, where the objective is to learn the time interval between two pulses.
The RSNN is excited with two input pulses, each with an amplitude of 1.0 and a time duration of 50ms. The RSNN is expected to generate an output response at a delay equal to the interval between the first and second pulse.
Figure 9 plots the MSE of proposed full-FORCE training with rate and TTFS coding compared to the FORCE training.
Results are reported for different interval matching tasks with intervals ranging from 0.1 to 2.1 in increment of 0.25.
%This time interval is chosen between 0.1 and 2.1 seconds with an increment of 0.25. Following the input pulses, the network is required to respond with an output pulse \ineq{f_{out}(t)} generated from a beta distribution with parameters \ineq{\alpha} and \ineq{\beta} = 4, time duration of 500ms,  and  peak amplitude of 1.5.  Results are shown for each interval matching task.
We observe that the MSE of full-FORCE training (both rate and TTFS coding) is lower than the FORCE.

\begin{figure}[h!]
	\centering
	%\vspace{-10pt}
	\centerline{\includegraphics[width=0.89\columnwidth]{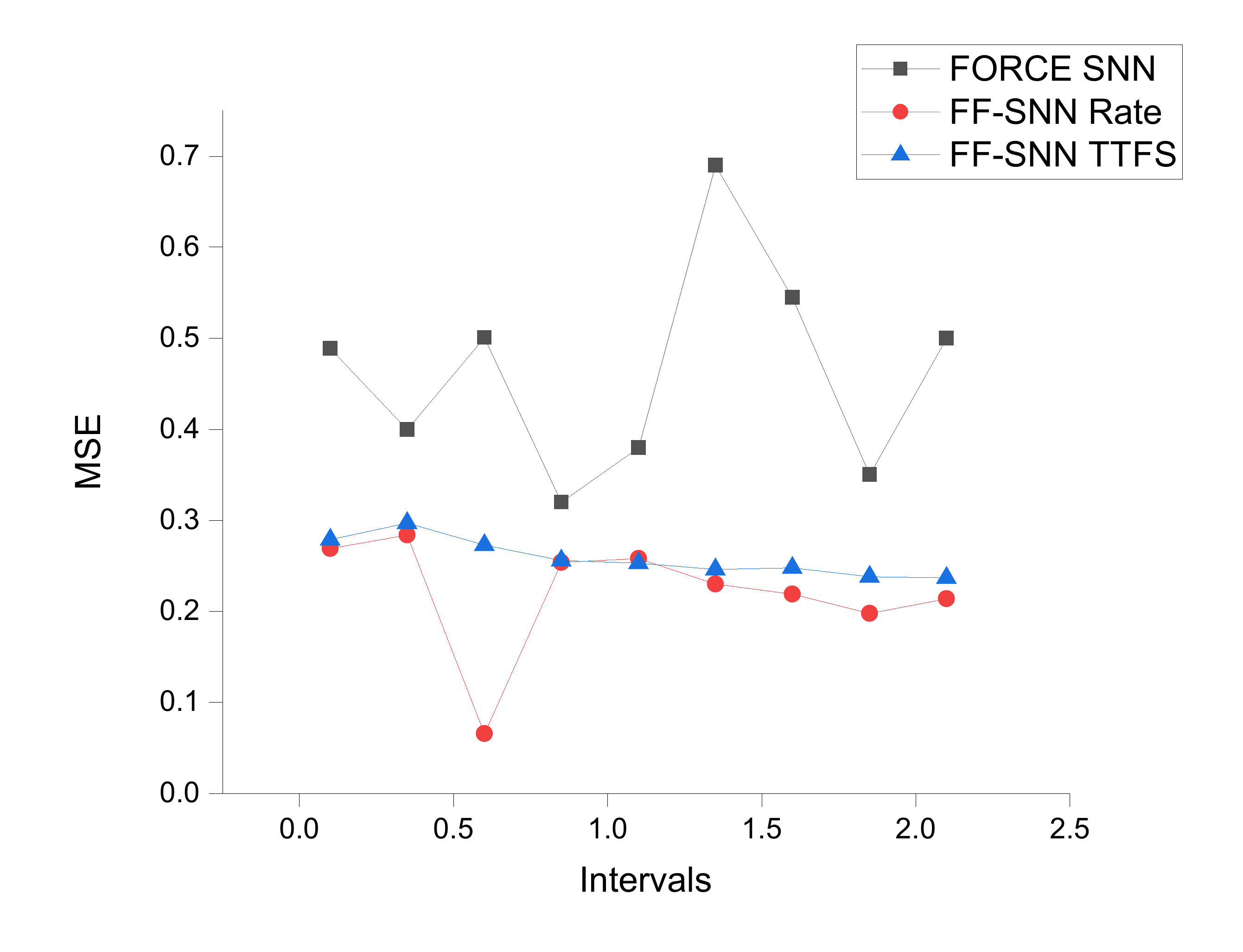}}
	%\vspace{-10pt}
	%\caption{An example of spiking neural network.}
	\caption{MSE for interval matching task with the time interval between two pulses ranging from 0.1 to 2.1 in increment of 0.25.}
	\label{fig:im}
\end{figure}

%\begin{itemize}
%\color{red}{\item 
  
%}     %\end{itemize}

%\section{Related Works}\label{sec:related_works}

\section{Conclusion}\label{sec:conclusions}

We propose a supervised training procedure, called full-FORCE for feedback-driven reservoir computing architecture with input, reservoir, and readout layers, each of which is designed using 
%recurrent spiking neural networks (RSNNs) with 
Leaky Integrate and Fire (LIF) neurons.
The objective is to model dynamical systems.
The full-FORCE training procedure introduces a second recurrently-connected reservoir network only during the training to provide hints for the target dynamics. 
%In this way, full-FORCE 
It then uses the recursive least square-based First-Order and Reduced Control Error (FORCE) algorithm to fit the activity of the reservoir and the readout to their corresponding target.
%Overall, it generates targets for both the reservoir and readout.
%Contrary to existing training approaches where only the output layer of an RSNN is learned, 
% The proposed training procedure consists of generating targets for both the recurrently-connected reservoir and the readout layer (i.e., for a full RSNN system), and using the recursive least square-based First-Order and Reduced Control Error (FORCE) algorithm to fit the activity of each layer to its target. 
We demonstrate the improved performance and noise robustness of the full-FORCE training to model \napps{} dynamical systems using LIF neurons with rate coding. 
For energy-efficient hardware implementation, an alternative time-to-first-spike (TTFS) coding scheme is also presented for the full-FORCE training. 
Compared to rate coding, full-FORCE with TTFS coding requires lower spike count and facilitates faster convergence to the target dynamics.%~\cite{valentian2019fully}.% with only marginally lower performance.

%The proposed training procedure is designed to learn dynamical systems that can be formulated with vector-valued functions as equations. To learn real-time dynamical systems, the continuous vectors can be trained as individual components of a function. We reserve this as our future work. We plan to conduct ablation experiments to assess the variability of error and spike rates in times of system failures and deficits.

% \section{Methods}\label{sec:methods}
% \input{sections/methods}

\section*{Acknowledgment}
This material is based upon work supported by the U.S. Department of Energy under Award Number DE-SC0022014 and by the National Science Foundation under Grant
Nos. CCF-1942697 and CCF-1937419.
% The preferred spelling of the word ``acknowledgment'' in America is without 
% an ``e'' after the ``g''. Avoid the stilted expression ``one of us (R. B. 
% G.) thanks $\ldots$''. Instead, try ``R. B. G. thanks$\ldots$''. Put sponsor 
% acknowledgments in the unnumbered footnote on the first page.

%\section*{References}
\bibliographystyle{IEEEtran}
\IEEEtriggeratref{42}
\bibliography{commands,disco,external}

\end{document}